\documentclass[10pt,twocolumn,letterpaper]{article}
\usepackage[pagenumbers]{cvpr}
\usepackage{multirow}
\usepackage{listings}
\usepackage{placeins}
\usepackage{needspace}
\graphicspath{{images/}}

\usepackage{amsmath,amsfonts,bm}

\def\eqref#1{equation~\ref{#1}}

\def\1{\bm{1}}

\DeclareMathAlphabet{\mathsfit}{\encodingdefault}{\sfdefault}{m}{sl}
\SetMathAlphabet{\mathsfit}{bold}{\encodingdefault}{\sfdefault}{bx}{n}

\newif\ifshowrevisions
\showrevisionsfalse

\DeclareRobustCommand{\rev}[1]{{\revisioncolor#1}}
\makeatletter
\@namedef{revisionbib@liu2026gdpo}{}
\@namedef{revisionbib@wang2025otcpo}{}
\@namedef{revisionbib@wang2025steca}{}
\@namedef{revisionbib@yuan2025agentr}{}
\newcommand{\revision@markedbibitem}[2][]{%
  \ifshowrevisions
    \ifcsname revisionbib@#2\endcsname
      \color{red}\hypersetup{citecolor=red,linkcolor=red,urlcolor=red}%
    \else
      \color{black}\hypersetup{citecolor=cvprblue,linkcolor=cvprblue,urlcolor=cvprblue}%
    \fi
  \fi
  \ifstrempty{#1}{\revision@originalbibitem{#2}}{\revision@originalbibitem[#1]{#2}}%
}
\AtBeginEnvironment{thebibliography}{%
  \let\revision@originalbibitem\bibitem
  \let\bibitem\revision@markedbibitem
}
\makeatother

\makeatletter
\@ifpackageloaded{lineno}{%
  \IfFormatAtLeastTF{2025-06-01}{%
    \RemoveFromHook{build/column/before}[lineno]
    \AtBeginDocument{%
      \def\@LN@makecol{%
        \@LN@orig@makecol
        \setbox\@outputbox\vbox{%
          \boxmaxdepth\@maxdepth
          \protected@write\@auxout{}{%
            \string\def\string\@LN@column{\if@firstcolumn1\else2\fi}%
          }%
          \box\@outputbox
        }%
      }%
    }%
  }{}%
}{}
\makeatother

\definecolor{cvprblue}{rgb}{0.21,0.49,0.74}
\usepackage[pagebackref,breaklinks,colorlinks,allcolors=cvprblue]{hyperref}
\def\confName{CVPR}
\def\confYear{2027}
\title{\rev{The Cost of Failure: Quantifying the Downstream Impact\\
of Error Histories in LLM Agents}}
\author{
  Jiahao Wang\textsuperscript{1,*}\quad
  Ruichuan An\textsuperscript{2,*}\quad
  Xinru Jiang\textsuperscript{3}\quad
  Sihan Yang\textsuperscript{2}\quad
  Hao Liang\textsuperscript{2}\\
  Bohan Zeng\textsuperscript{2}\quad
  YiFan Zhang\textsuperscript{4}\quad
  Aijun Yang\textsuperscript{1,\textdagger}\quad
  Xiaohua Wang\textsuperscript{1,\textdagger}\\[4pt]
  \small\textsuperscript{1}Xi'an Jiaotong University\quad
  \textsuperscript{2}Peking University\quad
  \textsuperscript{3}Northeastern University\\
  \small\textsuperscript{4}National Laboratory of Pattern Recognition,
  Institute of Automation, Chinese Academy of Sciences
}
\begin{document}
\maketitle
\begingroup
\renewcommand{\thefootnote}{\fnsymbol{footnote}}
\footnotetext[1]{Equal contribution.}
\footnotetext[2]{Co-corresponding authors.}
\endgroup
\begin{abstract}
When an action fails, an LLM agent retains the action text and its feedback in the context used for subsequent decisions. We study the downstream impact of this retained content as the \textit{Cost of Failure}. We define Error Token Density (ETD) as the fraction of an interaction history occupied by execution-failure content. On natural ALFWorld rollouts from Qwen3-8B, success is 40.1\% at zero ETD and 6.7\% when ETD is at least 40\%. Controlled continuations separate this association from the effect of retained failure content: with environment state and remaining action budget fixed, inserting failure turns lowers success by 1.9--3.8 percentage points more than token-matched neutral insertions. Deleting existing failure turns yields complementary gains. Failure-conditioned evaluations across planning, code execution, and visual tool use establish the broader relevance of the problem. These findings motivate training agents to reduce the introduction of execution failures. We test this principle with Tool-Feedback Advantage Shaping (TFAS), which combines separately normalized task-success and trajectory-level execution-validity advantages. Across all six model--environment settings, TFAS improves overall accuracy and achieves the lowest ETD among the compared policies, reducing ETD by 26.5--67.7\% relative to outcome-only RL. These measurements connect the intervention to lower failure-context density.
\end{abstract}

\section{Introduction}

\begin{figure*}[t]
    \centering
    \includegraphics[width=\textwidth]{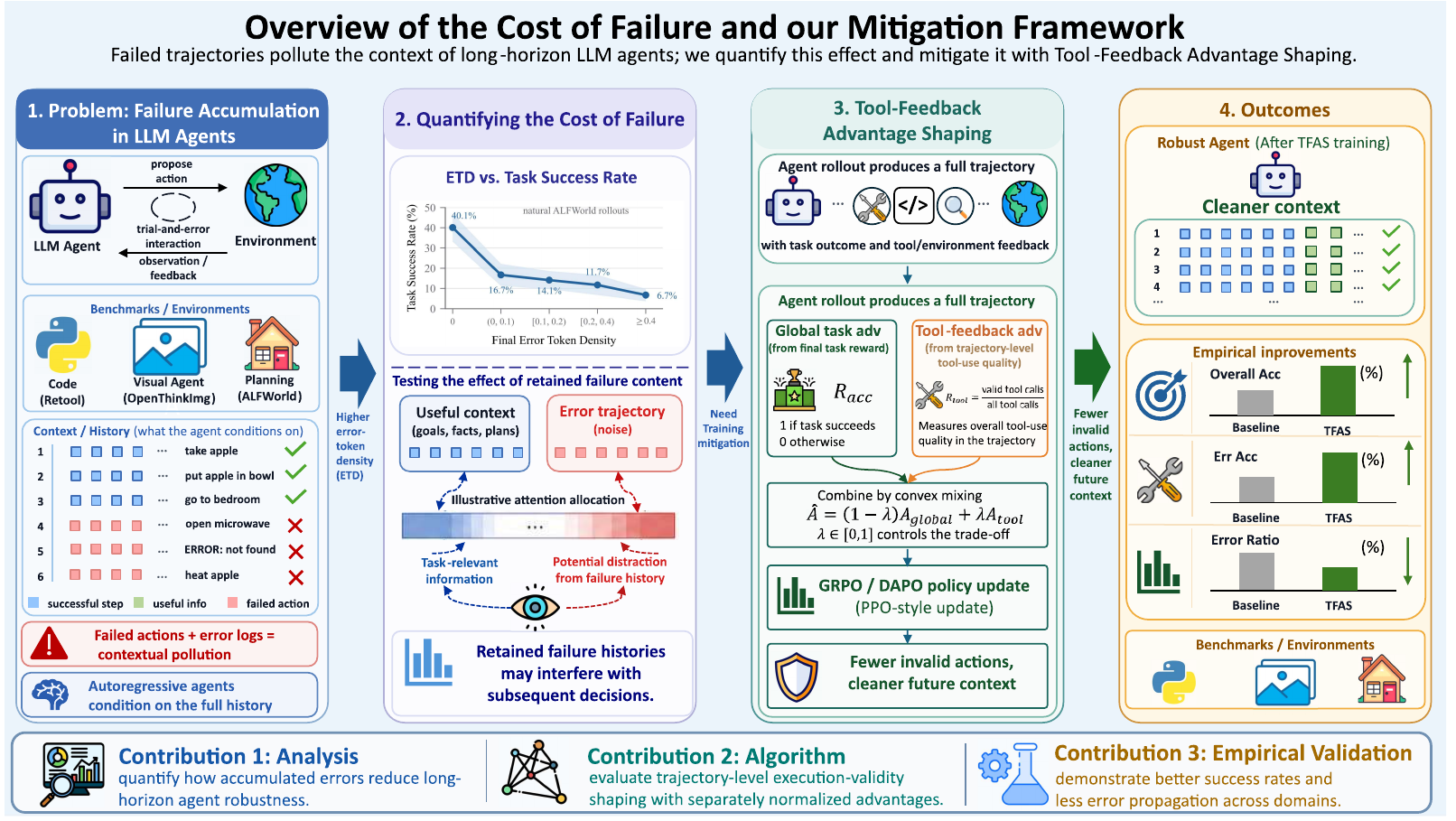}
    \caption{\textbf{Overview of the Cost of Failure and TFAS.} Failed actions and their feedback persist in the context for later decisions. ETD measures this content, while controlled interventions test its downstream effects. TFAS combines task-success and trajectory-level execution-validity advantages to favor trajectories with fewer execution failures.}
    \label{fig:overview}
\end{figure*}

LLM agents solve interactive tasks by alternating between actions and environmental feedback~\cite{yao2023react,schick2023toolformer,shridhar2021alfworld}. When an action fails, the feedback can reveal a missing prerequisite or an unavailable operation and help the agent revise its plan. Self-correction methods such as Reflexion~\cite{shinn2023reflexion} and Self-Refine~\cite{madaan2023selfrefine} exploit this information. The same interaction also becomes part of the prompt for subsequent decisions. An execution failure therefore has a continuing textual footprint even when it leaves the environment unchanged.

This persistence raises a question about agent reliability: \emph{how does retaining failed interactions affect what an agent does next?} Long-context studies show that available context is not always used reliably~\cite{liu2024lost}, and irrelevant content can distract language models~\cite{shi2023distracted}. Repeated invalid calls and error feedback can accumulate alongside the task goal and useful observations. We call the downstream performance cost of retaining this content the \textit{Cost of Failure}. Understanding it requires separating the presence of errors from the effect of their textual history: difficult tasks may both elicit more errors and have lower success rates.

We study that distinction through complementary analyses. We define Error Token Density (ETD) as the fraction of a trajectory history occupied by failed actions and their associated feedback. On natural ALFWorld rollouts from Qwen3-8B~\cite{qwen3technicalreport}, success is 40.1\% at zero ETD and 6.7\% when ETD is at least 0.4. To test the effect of the history itself, we then resume matched prefixes from the same environment state and remaining action budget. Inserting invalid-action/error-feedback turns reduces success by 1.9--3.8 percentage points more than inserting token-matched neutral turns. Removing existing failure turns also improves continuation success. These interventions identify a cost of retained failure content in the studied ALFWorld setting, beyond the cost of simply extending the context. Failure-conditioned evaluations in ReTool, OpenThinkImg, and ALFWorld show that errors accompany lower success across text and vision-language agents, including advanced closed-source models. The cross-domain comparison establishes the prevalence of the association; the matched interventions isolate a causal effect within ALFWorld.

The analysis suggests a training principle: \emph{reduce avoidable execution failures before their feedback becomes part of future context}. Final task reward alone leaves trajectories with the same outcome indistinguishable even when their execution histories differ. We test whether adding this distinction helps by using Tool-Feedback Advantage Shaping (TFAS). TFAS computes the fraction of executable tool calls in a completed trajectory, normalizes this signal separately from task success, and mixes the two advantages for policy optimization. Its optimization form follows the general idea of reward-decoupled normalization~\cite{liu2026gdpo}; the choice of execution validity is motivated by the failure-context analysis. The resulting preference is between trajectories. All response tokens within a trajectory share the same mixed advantage, so the method does not assign separate credit to individual successful and failed actions. The training results show that this intervention improves overall accuracy across six settings spanning ReTool and ALFWorld with Qwen3-4B/8B, and OpenThinkImg with Qwen3-VL-4B/8B. Relative to outcome-only RL, Qwen3-8B improves from 55.1\% to 58.4\% on ReTool and from 31.2\% to 41.2\% on ALFWorld. The reported Err Proportion is also lower than under outcome-only RL in all six settings. These results establish the value of the training signal for task completion. In addition, TFAS has the lowest ETD among all four policies in every setting. Relative to outcome-only RL, ETD falls by 26.5--67.7\%; on ALFWorld with Qwen3-8B, it decreases from 8.7\% to 2.8\%. Reward-level tool-validity shaping increases ETD in both ReTool settings and ALFWorld-4B, showing that the same auxiliary signal does not uniformly reduce failure-context density when added at the reward level. These results connect the training intervention to the context property identified by the analysis.

Overall, our contributions are:
\begin{itemize}
    \item We characterize retained execution failures with ETD and cross-domain failure-conditioned evaluations, distinguishing failure exposure from success after errors.
    \item We use matched context insertions and deletions to identify a downstream cost of retained failure content in ALFWorld, including an insertion comparison that controls for added token count.
    \item We train agents to reduce execution failures with TFAS, a trajectory-level execution-validity signal. TFAS improves overall accuracy and yields the lowest measured ETD in all six evaluated settings.
\end{itemize}

\Needspace{6\baselineskip}
\section{Related Work}

\noindent\textbf{Agent failures and correction.}
ReAct~\cite{yao2023react} and Toolformer~\cite{schick2023toolformer} establish reasoning--action and tool-use paradigms, and interactive benchmarks such as ALFWorld~\cite{shridhar2021alfworld}, WebShop~\cite{yao2022webshop}, WebArena~\cite{zhou2024webarena}, VisualWebArena~\cite{koh2024visualwebarena}, and AgentGym~\cite{xi2024agentgym} expose the demands of multi-step decision making. Reflexion~\cite{shinn2023reflexion} and Self-Refine~\cite{madaan2023selfrefine} use feedback to revise behavior, while AgentBoard~\cite{ma2024agentboard} and AgentErrorBench/AgentDebug~\cite{zhu2025wherellmagentsfail} analyze partial progress and failure modes. STeCa~\cite{wang2025steca} and Agent-R~\cite{yuan2025agentr} train agents to correct deviations through step-level calibration and revision trajectories. These works study how to recognize and correct errors; we study the effect of retaining failure content in the autoregressive history.

\noindent\textbf{Context use and retained errors.}
Long context is not always used effectively~\cite{liu2024lost}, and irrelevant content can distract language models~\cite{shi2023distracted}. Failed actions and their feedback add such context during interactive execution, persisting alongside useful observations and task progress as evidence for later decisions. ETD measures this accumulated failure content, and matched interventions test its effect with environment state and action budget fixed. The evidence concerns behavioral sensitivity to history and leaves the attention-level mechanism unidentified.

\noindent\textbf{Training signals for tool-using agents.}
PPO-style optimization~\cite{schulman2017ppo,schulman2015gae}, outcome rewards, and preference learning~\cite{ouyang2022training} support agent training in tool use, web interaction, and multi-environment settings~\cite{feng2025retoolreinforcementlearningstrategic,qi2025webrl,chen2025loop,luo2025agentlightning,xi2025agentgymrl}. RAGEN~\cite{wang2025ragen} analyzes multi-turn training instability, rollback systems revise failed histories at inference time~\cite{li2025garollback,ma2025dover,li2025agentgit}, and reward shaping~\cite{xie2024text2reward} and process supervision~\cite{lightman2023lets} offer additional signals. TFAS uses execution validity already returned by tools or environments, aggregated at the trajectory level. Its optimization form is closely related to GDPO~\cite{liu2026gdpo}, which normalizes each reward within a prompt group before aggregation; we use this construction to test a signal selected from the failure-context analysis. OTC-PO~\cite{wang2025otcpo} modulates task reward with a call-count-based reward to improve tool-call efficiency, whereas execution validity measures whether an invoked operation can be executed.

\section{Analysis}
\label{sec:cost_of_failure}

A failed action leaves its text and error feedback in the history that conditions every later decision. This section asks whether that retained content itself lowers subsequent performance. Observational statistics alone cannot answer this question, because harder tasks both elicit more errors and succeed less often. We address it with natural-rollout statistics, matched context interventions, and cross-domain failure-conditioned evaluations.

\begin{figure}[t]
    \centering
    \includegraphics[width=\columnwidth]{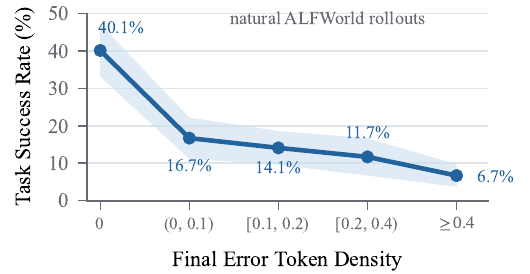}
    \caption{\textbf{Task success rate across final-ETD buckets for Qwen3-8B on natural ALFWorld rollouts.} Higher Error Token Density corresponds to a larger fraction of the accumulated context being occupied by invalid actions and error feedback, and is associated with lower task success.}
    \label{fig:cost_of_failure_curve}
\end{figure}

\begin{table}[t]
    \centering
    \caption{\textbf{Controlled failure-context interventions on ALFWorld.} Values are changes in success rate (percentage points). Each insertion or deletion is compared with a matched trajectory with an unmodified history.}
    \label{tab:controlled_interventions}
    \small
\begin{tabular}{ll rr}
\toprule
\textbf{Intervention} & $k$ & $\rho=0.25$ & $\rho=0.50$ \\
\midrule
\multirow{3}{*}{Error insertion ($E_k$)}
& 1 & $-4.7$ & $-4.3$ \\
& 2 & $-5.7$ & $-6.0$ \\
& 3 & $-8.2$ & $-8.8$ \\
\midrule
\multirow{3}{*}{Neutral insertion ($N_k$)}
& 1 & $-0.9$ & $-2.3$ \\
& 2 & $-3.8$ & $-3.0$ \\
& 3 & $-5.6$ & $-5.9$ \\
\midrule
\multirow{3}{*}{Failure deletion ($D_k$)}
& 1 & $+2.5$  & $+2.0$  \\
& 2 & $+20.0$ & $+13.6$  \\
& 3 & -- & -- \\
\bottomrule
\end{tabular}
\end{table}

\begin{table*}[t]
\centering
\caption{\textbf{Failure-conditioned performance across three agent domains.} SR: success rate; Prop.: trajectory proportion; Gap: error-free minus error-containing SR. The split describes the association between observed execution errors and task success.}
\label{tab:failure_conditioned_domains}
\resizebox{\textwidth}{!}{%
\begin{tabular}{ll cccccc}
\toprule
\multirow{2}{*}{\textbf{Environment}} & \multirow{2}{*}{\textbf{Model}} & \textbf{Overall} & \multicolumn{2}{c}{\textbf{With Error}} & \multicolumn{2}{c}{\textbf{Without Error}} & \textbf{Gap} \\
\cmidrule(lr){3-3} \cmidrule(lr){4-5} \cmidrule(lr){6-7} \cmidrule(lr){8-8}
& & SR (\%) & SR (\%) & Prop. (\%) & SR (\%) & Prop. (\%) & $\Delta$ SR (\%) \\
\midrule
\multirow{5}{*}{ReTool} 
& GPT-5~\cite{openai2025gpt5}             & 65.5 & 61.5 & 19.1 & 90.9 & 80.9 & 29.4 \\
& Claude-Sonnet-4~\cite{anthropic2025claude4}   & 52.1 & 40.2 & 46.7 & 62.5 & 53.3 & 22.3 \\
& Gemini-3-Flash~\cite{google2025gemini3flash}    & 46.7 & 28.5 & 23.3 & 56.2 & 76.7 & 27.7 \\
& DeepSeek-V3.2~\cite{deepseekai2025deepseekv32}     & 76.5 & 50.0 & 1.2 & 91.6 & 98.8 & \textbf{41.6} \\
& Qwen3-8B~\cite{qwen3technicalreport}          & 51.3 & 49.3 & 30.0 & 68.3 & 70.0 & 19.0 \\
\midrule
\multirow{5}{*}{OpenThinkImg}
& GPT-5~\cite{openai2025gpt5}             & 48.9 & 19.8 & 16.0 & 71.4 & 84.0 & 51.6 \\
& Claude-Sonnet-4~\cite{anthropic2025claude4}   & 79.9 & 39.5 & 25.2 & 60.5 & 74.8 & 21.0 \\
& Gemini-3-Flash~\cite{google2025gemini3flash}    & 82.7 & 38.9 & 28.1 & 61.1 & 71.9 & 22.2 \\
& Doubao-Seed-1.6~\cite{bytedance2025seed16}     & 73.1 & 10.0 & 2.3 & 75.7 & 97.7 & \textbf{65.7} \\
& Qwen3-VL-8B-Instruct~\cite{bai2025qwen3vl}      & 48.8 & 16.2  & 1.4 & 73.8 & 98.6 & 57.6 \\
\midrule
\multirow{5}{*}{ALFWorld}
& GPT-5.1~\cite{openai2025gpt51}             & 76.4 & 33.3 & 9.4 & 80.5 & 90.6 & 47.2 \\
& Claude-Sonnet-4~\cite{anthropic2025claude4}      & 39.3 & 27.8 & 77.1 & 78.1 & 22.9 & \textbf{50.3} \\
& Gemini-3-Flash~\cite{google2025gemini3flash}       & 59.3 & 42.9 & 5.0 & 60.2 & 95.0 & 17.3 \\
& DeepSeek-V3.2~\cite{deepseekai2025deepseekv32}     & 36.4 & 20.6 & 24.3 & 41.5 & 75.7 & 20.9 \\
& Qwen3-8B~\cite{qwen3technicalreport}        & 31.4 & 8.2 & 27.0 & 40.1 & 73.0 & 31.9 \\
\bottomrule
\end{tabular}%
}
\begin{flushleft}
\footnotesize \textit{Note.} Prop. is computed over trajectories with at least one tool invocation; trajectories where the model does not call a tool are included in Overall SR but excluded from the error-conditioned proportion split.
\end{flushleft}

\end{table*}

\subsection{Failure Histories as Autoregressive Context}
\label{subsec:error_trajectory_formalization}

Before action $a_t$, an agent has a textual interaction history
\[
\tau_t=(o_1,a_1,o_2,\ldots,a_{t-1},o_t),
\]
where $o_t$ is the latest observation. The policy generates the action token by token,
\[
P(a_t\mid\tau_t)=\prod_{i=1}^{|a_t|}P(x_{t,i}\mid\tau_t,x_{t,<i}).
\]
An invalid call may leave the environment state unchanged while extending $\tau_t$ with the attempted action and its error feedback. Subsequent actions are thus conditioned on a different history even though the physical or tool state is the same. This difference makes it possible to isolate the effect of retained failure content.

\subsection{Measuring Failure Exposure with ETD}
\label{subsec:error_token_density}

We label observable execution failures: malformed calls, exceptions, and rejected environment actions. For a completed trajectory, let $\tau_{err}$ denote the tokens in these failed actions and their associated error feedback. Error Token Density is
\[
\operatorname{ETD}(\tau)=\frac{|\tau_{err}|}{|\tau|},
\]
where $|\tau|$ counts all tokens in the accumulated interaction history. ETD measures the share of context occupied by execution failures, incorporating the amount of text returned by those failures as well as their frequency.

We compute final ETD on natural Qwen3-8B~\cite{qwen3technicalreport} rollouts in ALFWorld. Figure~\ref{fig:cost_of_failure_curve} shows a large difference between trajectories with no error tokens and those containing them. Success is 40.1\% at zero ETD, ranges from 11.7\% to 16.7\% for nonzero ETD below 0.4, and falls to 6.7\% when ETD is at least 0.4. This is an association in completed rollouts: task difficulty, early completion, and repeated failed attempts can affect both final ETD and success. We next intervene on the history to isolate an effect of retained content.

\subsection{Controlled Failure-Context Interventions}
\label{subsec:controlled_interventions}

At normalized trajectory anchors $\rho\in\{0.25,0.5\}$, we branch from the same unfinished prefix and environment state with the same remaining action budget. Error insertion $E_k$ adds $k\in\{1,2,3\}$ state-preserving invalid-action/error-feedback turns. Neutral insertion $N_k$ adds the same number of turns, with token counts matched to the corresponding error content. Failure deletion $D_k$ removes the most recent $k$ complete failure turns from an eligible error-containing prefix. Each branch is compared with continuation from its unmodified history.

Table~\ref{tab:controlled_interventions} shows that inserting one failure turn lowers success by 4.3--4.7 percentage points; three turns lower it by 8.2--8.8 points. Neutral insertions also reduce success, showing that added context has a cost. Across matched insertion settings, the error content causes an additional reduction of 1.9--3.8 points. Because the environment state, future action budget, turn count, and added token count are controlled, this comparison identifies an effect of the inserted failure content beyond context growth in this setting.

Deleting one existing failure turn improves success by 2.0--2.5 points, and deleting two improves it by 13.6--20.0 points. Deletion removes both content and tokens, so its gains combine these effects, and the $k{=}1$ and $k{=}2$ gains are measured on different sets of eligible prefixes. Together, insertion and deletion show that keeping failure history can impede later decisions, and that changing this history can alter task completion without changing the environment state at the branching point. Appendix~\ref{apdx:intervention_repeats} reports repeated interventions.

\begin{table*}[!t]
\centering
\caption{\textbf{Main Experimental Results across Three Environments.} We report the Overall Accuracy (Overall Acc), the proportion of trajectories that contain at least one error (Err Proportion), and the task success rate for trajectories that contain at least one error step (Err Acc). The untrained base model is included as the baseline. The best results are highlighted in \textbf{bold}. TFAS achieves the best overall accuracy in all settings, demonstrating the effectiveness of our method. \rev{ETD is the mean per-episode error-token share, including zero-error episodes.}}
\label{tab:main_results}
\small
\setlength{\tabcolsep}{3pt}
\begin{tabular*}{\textwidth}{@{\extracolsep{\fill}}lllrrrr@{}}
\toprule
\textbf{Environment} & \textbf{Model} & \textbf{Reward Strategy} & \shortstack{\textbf{Overall Acc}\\(\%)$\uparrow$} & \shortstack{\textbf{Err Proportion}\\(\%)$\downarrow$} & \shortstack{\textbf{Err Acc}\\(\%)$\uparrow$} & \rev{\shortstack{\textbf{ETD}\\(\%)$\downarrow$}} \\
\midrule
\multirow{8}{*}{\textbf{ReTool}}
& \multirow{4}{*}{Qwen3-4B~\cite{qwen3technicalreport}}
& Base Model (No Training) & 47.9 & 40.3 & 23.7 & \rev{1.6} \\
& & Original Outcome Reward & 50.4 & 41.4 & 22.7 & \rev{1.8} \\
& & + Tool Valid Reward     & 52.5 & 38.8 & 19.1 & \rev{1.9} \\
& & + \textbf{TFAS (Ours)} & \textbf{54.7} & \textbf{34.6} & \textbf{33.3} & \rev{\textbf{0.9}} \\
\cmidrule{2-7}
& \multirow{4}{*}{Qwen3-8B~\cite{qwen3technicalreport}}
& Base Model (No Training) & 51.3 & 30.0 & 49.3 & \rev{0.7} \\
& & Original Outcome Reward & 55.1 & 38.4 & 26.2 & \rev{0.8} \\
& & + Tool Valid Reward     & 56.2 & 31.2 & 25.4 & \rev{1.0} \\
& & + \textbf{TFAS (Ours)} & \textbf{58.4} & \textbf{28.5} & \textbf{53.6} & \rev{\textbf{0.6}} \\
\midrule
\multirow{8}{*}{\textbf{OpenThinkImg}}
& \multirow{4}{*}{Qwen3-VL-4B~\cite{bai2025qwen3vl}}
& Base Model (No Training) & 42.4 & 7.8 & 15.5 & \rev{2.6} \\
& & Original Outcome Reward & 44.3 & 7.1 & \textbf{19.4} & \rev{3.4} \\
& & + Tool Valid Reward     & 46.5 & 5.7 & 9.5 & \rev{2.0} \\
& & + \textbf{TFAS (Ours)} & \textbf{49.3} & \textbf{3.8} & 10.7 & \rev{\textbf{1.5}} \\
\cmidrule{2-7}
& \multirow{4}{*}{Qwen3-VL-8B~\cite{bai2025qwen3vl}}
& Base Model (No Training) & 48.8 & \textbf{1.4} & 16.2 & \rev{1.0} \\
& & Original Outcome Reward & 49.2 & 12.6 & 18.4 & \rev{1.1} \\
& & + Tool Valid Reward     & 49.5 & 2.7 & 16.7 & \rev{1.0} \\
& & + \textbf{TFAS (Ours)} & \textbf{52.9} & 2.2 & \textbf{20.9} & \rev{\textbf{0.8}} \\
\midrule
\multirow{8}{*}{\textbf{ALFWorld}}
& \multirow{4}{*}{Qwen3-4B~\cite{qwen3technicalreport}}
& Base Model (No Training) & 17.9 & 83.6 & 10.3 & \rev{18.1} \\
& & Original Outcome Reward & 17.8 & 84.2 & 12.3 & \rev{23.7} \\
& & + Tool Valid Reward     & 20.9 & 66.7 & 22.1 & \rev{36.1} \\
& & + \textbf{TFAS (Ours)} & \textbf{24.3} & \textbf{64.1} & \textbf{31.8} & \rev{\textbf{10.1}} \\
\cmidrule{2-7}
& \multirow{4}{*}{Qwen3-8B~\cite{qwen3technicalreport}}
& Base Model (No Training) & 31.4 & 27.0 & 8.2 & \rev{6.7} \\
& & Original Outcome Reward & 31.2 & 18.3 & 21.0 & \rev{8.7} \\
& & + Tool Valid Reward     & 33.6 & 18.0 & 35.7 & \rev{8.0} \\
& & + \textbf{TFAS (Ours)} & \textbf{41.2} & \textbf{15.8} & \textbf{52.4} & \rev{\textbf{2.8}} \\
\bottomrule
\end{tabular*}

\end{table*}

\subsection{Failure-Conditioned Performance Across Domains}
\label{subsec:task_level_evidence}

We broaden the observational analysis to ReTool~\cite{feng2025retoolreinforcementlearningstrategic} for code-interpreter-assisted mathematics, OpenThinkImg~\cite{su2025openthinkimg} for visual tool reasoning, and ALFWorld~\cite{shridhar2021alfworld} for embodied planning. We divide evaluated tool-using trajectories according to whether they contain at least one execution error. The evaluated models include GPT-5/5.1~\cite{openai2025gpt5,openai2025gpt51}, Claude-Sonnet-4~\cite{anthropic2025claude4}, Gemini-3-Flash~\cite{google2025gemini3flash}, Doubao-Seed-1.6~\cite{bytedance2025seed16}, DeepSeek-V3.2~\cite{deepseekai2025deepseekv32}, Qwen3~\cite{qwen3technicalreport}, and Qwen3-VL~\cite{bai2025qwen3vl}.

Table~\ref{tab:failure_conditioned_domains} reports lower success for error-containing trajectories in all listed settings. For example, GPT-5 on ReTool falls from 90.9\% on error-free trajectories to 61.5\% on trajectories with errors. In ALFWorld, the corresponding values are 40.1\% and 8.2\% for Qwen3-8B, and 80.5\% and 33.3\% for GPT-5.1. OpenThinkImg illustrates why error frequency and error-conditioned performance should be kept separate: Qwen3-VL-8B-Instruct has errors in only 1.4\% of evaluated tool-using trajectories, yet those trajectories succeed at 16.2\%, compared with 73.8\% for the error-free subset.

These comparisons show that the failure-conditioned gap extends across domains and model families. They are observational and complement the controlled interventions, which establish that retained history itself can affect continuation.

\begin{figure*}[!t]
    \centering
    \includegraphics[width=0.95\textwidth]{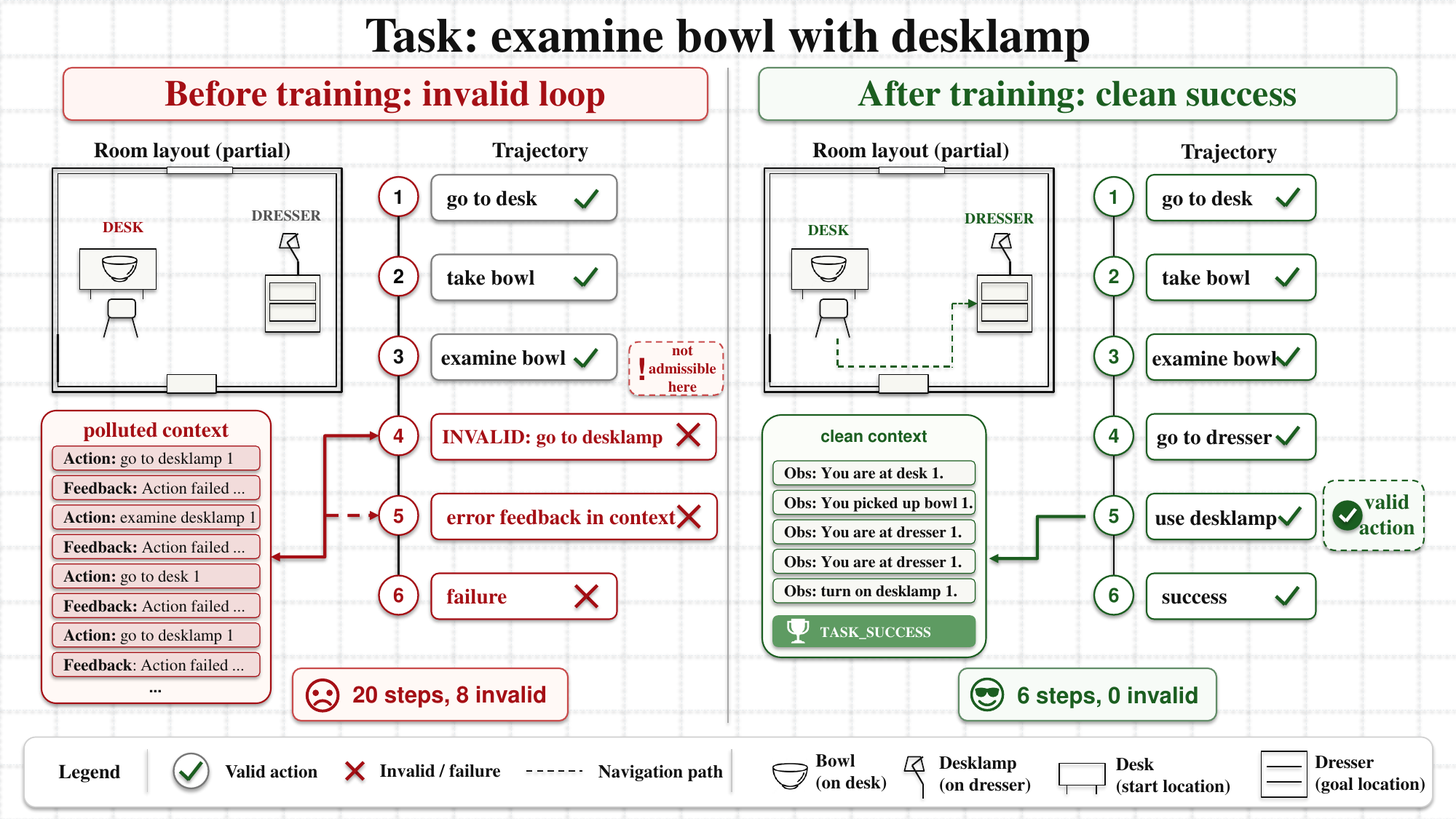}
    \caption{\textbf{Qualitative ALFWorld case comparing behavior before and after TFAS training.} The baseline policy enters an invalid-action loop, while the trained policy follows executable actions and completes the task without invalid steps.}
    \label{fig:qualitative_tool_failure_recovery}
\end{figure*}

\section{Tool-Feedback Advantage Shaping}
\label{sec:method}

Section~\ref{sec:cost_of_failure} shows that retained failure content reduces continuation success. Outcome-only training does not account for this cost: trajectories with the same final result receive the same reward regardless of how many failures they introduce. Tool-Feedback Advantage Shaping (TFAS) adds an execution-validity signal to a GRPO/DAPO update~\cite{shao2024deepseekmath,yu2025dapo}, so that the policy is also rewarded for keeping failures out of its history. The signal is read directly from tool and environment feedback and requires neither a learned reward model nor step-level annotation.

\subsection{A Trajectory-Level Execution Signal}
\label{subsec:tool_feedback_advantage}

Let $c_t\in\{0,1\}$ indicate whether step $t$ invokes a tool and $v_t\in\{0,1\}$ indicate whether that invocation is valid and executable. For a completed trajectory $\tau$, define
\[
R^{tool}(\tau)=
\begin{cases}
\displaystyle\frac{\sum_{t=1}^{T}c_tv_t}{\sum_{t=1}^{T}c_t}, & \sum_{t=1}^{T}c_t>0,\\[3pt]
0, & \sum_{t=1}^{T}c_t=0.
\end{cases}
\]
This score is available during rollout at no extra cost: tools and environments already report whether each call executes. The same definition therefore covers code errors, visual-tool invocation failures, and rejected environment actions.

For trajectories sampled from the same prompt group $\mathcal G$, the tool-feedback advantage is
\[
A^{tool}(\tau)=\frac{R^{tool}(\tau)-\mu_{\mathcal G}^{tool}}
{\sigma_{\mathcal G}^{tool}+\epsilon},
\]
where $\mu_{\mathcal G}^{tool}$ and $\sigma_{\mathcal G}^{tool}$ are the group mean and standard deviation. We compute $A^{acc}$ in the same way from the binary task-success reward. Normalizing within a group compares trajectories that attempt the same task, so the advantage reflects how cleanly a trajectory executes relative to its peers. The tool advantage also remains informative when the outcome signal is not: if every sampled trajectory in a group succeeds, or every one fails, $A^{acc}$ is zero for all of them, while differences in execution validity still rank them.

\subsection{Mixing Advantages for Policy Optimization}
\label{subsec:mixed_advantage_objective}

We broadcast each trajectory's two scalar advantages to its response tokens and form
\begin{equation}
\label{eq:mixed_advantage}
\hat A_t=(1-\lambda)A_t^{acc}+\lambda A_t^{tool}.
\end{equation}
Here $t$ indexes a response token in the policy update, and $\lambda\in[0,1]$ sets the tool-signal weight; $\lambda=0$ recovers outcome-only training. The mixed advantage enters the usual clipped surrogate,
\[
\mathcal L(\theta)=\mathbb E_t\!\left[
\min\!\left(r_t(\theta)\hat A_t,
\operatorname{clip}(r_t(\theta),1-\epsilon,1+\epsilon)\hat A_t\right)
\right],
\]
where $r_t$ is the current-to-behavior policy probability ratio.

Mixing after normalization has two benefits. First, the two advantages have a comparable scale within every group, so $\lambda$ directly sets the balance between task success and execution validity. When the two rewards are combined before normalization, as in the Tool Valid Reward baseline, their relative influence depends on their within-group variances and changes from prompt to prompt. Second, TFAS changes only the advantage: it reuses the same rollouts, adds no model or environment interaction, and keeps task success in the objective for any $\lambda<1$. Section~\ref{sec:lambda_ablation} studies the choice of $\lambda$.

\section{Experiment}
\label{sec:experiment}

\begin{figure*}[!t]
    \centering
    \begin{minipage}{0.495\textwidth}
        \centering
        \includegraphics[width=\linewidth]{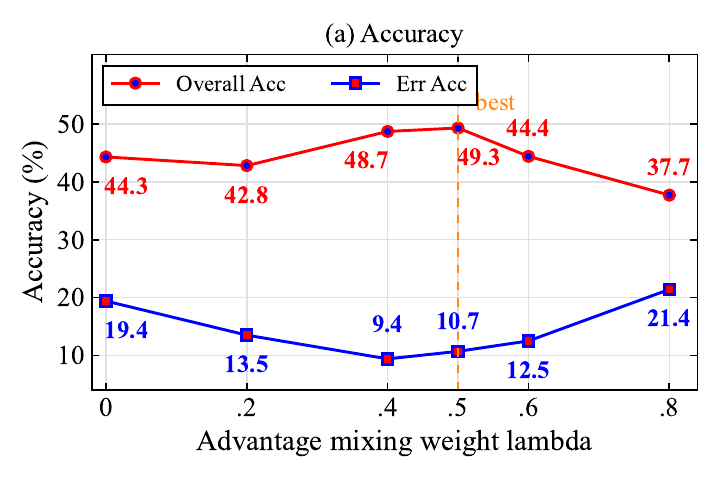}
    \end{minipage}
    \hfill
    \begin{minipage}{0.495\textwidth}
        \centering
        \includegraphics[width=\linewidth]{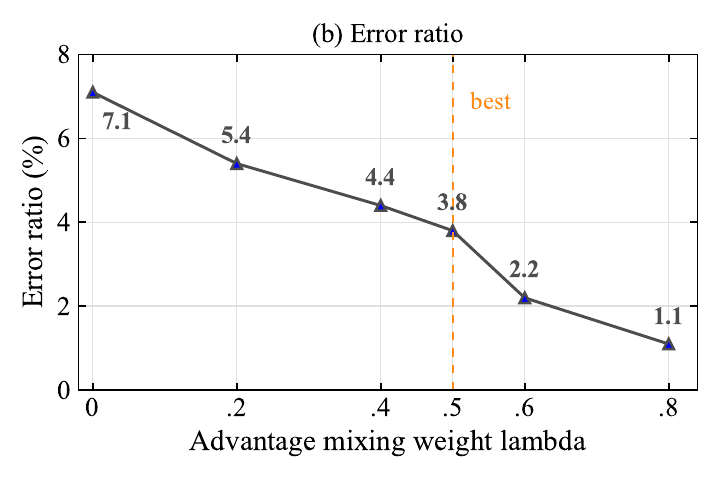}
    \end{minipage}
    \caption{\textbf{Sensitivity analysis of the advantage mixing weight $\lambda$ on OpenThinkImg with Qwen3-VL-4B.} (a) Overall Acc under different mixing weights. (b) Err Ratio under the same settings. A moderate $\lambda$ provides the best balance between task success and error reduction.}
    \label{fig:lambda_sensitivity}
\end{figure*}

We describe the setup, compare TFAS with outcome-only and tool-valid reward baselines, and analyze task success, error-containing trajectories, and advantage mixing weights.

\subsection{Experimental Setup}
\noindent \textbf{Environments and Datasets.} We test generalization across state spaces and error modalities in three long-horizon interactive environments. In \textbf{ReTool}~\cite{feng2025retoolreinforcementlearningstrategic}, agents interleave natural-language reasoning and executable Python to solve competition-style math problems. In \textbf{OpenThinkImg}~\cite{su2025openthinkimg}, VLM agents solve chart reasoning tasks by adaptively invoking external vision tools, including OCR and image zooming. In \textbf{ALFWorld}~\cite{shridhar2021alfworld}, a text-based embodied simulation, multi-step household tasks generate exploration-driven trial-and-error noise. Together, they test mitigation of contextual interference across code-tool reasoning, visual-linguistic, and physical domains.

\noindent \textbf{Baselines.} We evaluate effectiveness and generalization across text-only and vision-language modalities and 4B/8B scales using Qwen3-4B/8B~\cite{qwen3technicalreport} and Qwen3-VL-4B/8B~\cite{bai2025qwen3vl} as baseline models.

\noindent \textbf{Reward Strategies.} We compare three reinforcement learning strategies. \textbf{Original Outcome Reward (Baseline)} uses only the final accuracy reward (acc reward) upon task success. \textbf{Tool Validation Reward} adds a trajectory-level tool-valid reward, the proportion of correct tool-use steps in the completed rollout. \textbf{Tool-Feedback Advantage Shaping (TFAS, Ours)} converts this same signal into a separate advantage and mixes it with the accuracy advantage through Equation~\ref{eq:mixed_advantage}. Both tool-feedback strategies use the same tool-signal weight.




\subsection{Experiment Results}
\label{subsec:main_results}
Table~\ref{tab:main_results} shows consistent findings across the three environments.

\noindent \textbf{Higher Overall Acc.} Our TFAS achieves the highest Overall Acc in all six model-environment settings, over the base model, outcome-only RL, and tool-valid reward baselines. For example, it improves Qwen3-8B~\cite{qwen3technicalreport} from 51.3\% to 58.4\% on ReTool and from 31.4\% to 41.2\% on ALFWorld. This confirms that TFAS improves final task completion rather than only changing error statistics.

\noindent \textbf{Lower Err Proportion.} Our method usually reduces the proportion of error-containing trajectories. It obtains the lowest Err Proportion in ReTool and ALFWorld for both model sizes. In OpenThinkImg, for Qwen3-VL-8B~\cite{bai2025qwen3vl}, it also sharply reduces the error proportion compared with outcome-only RL, from 12.6\% to 2.2\%. This suggests that TFAS generally suppresses the formation of polluted contexts before they accumulate.

\noindent \textbf{Stronger Err Acc.} Error-conditioned accuracy measures the task-success rate after at least one error has already entered the trajectory. TFAS gives the best Err Acc in ReTool and ALFWorld, reaching 53.6\% on ReTool with Qwen3-8B~\cite{qwen3technicalreport} and 52.4\% on ALFWorld with Qwen3-8B~\cite{qwen3technicalreport}. Together with the Err Proportion reductions observed in many settings, this suggests that the method reduces harmful error injection while also making the remaining error-containing trajectories less likely to collapse.

\noindent \textbf{Tool Feedback Matters.} Our results suggest that tool-failure feedback is more effective when used beyond a plain reward bonus. Outcome-only RL often provides limited gains, while Tool Valid Reward sometimes lowers Err Acc, such as in ReTool and OpenThinkImg with Qwen3-VL-4B~\cite{bai2025qwen3vl}. Mixing tool feedback at the advantage level gives a clearer optimization signal than either ignoring failures or adding a shallow validity reward. Thus, the benefit comes from how tool feedback is injected into optimization, not merely from adding an auxiliary reward.

\noindent \textbf{Broad Applicability.} The same pattern holds across different tasks, modalities, and model sizes: text-only agents on ReTool and ALFWorld, VLM agents on OpenThinkImg, and both 4B and 8B scales. This suggests that TFAS addresses a general failure mode of agent training, and failure-induced context pollution is a cross-domain agent-training problem.

\begin{figure}[t]
    \centering
    \includegraphics[width=\columnwidth]{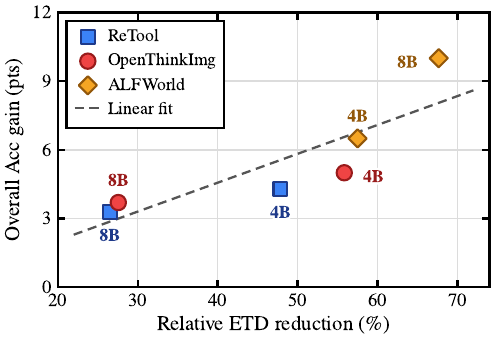}
    \caption{\rev{\textbf{Accuracy gain versus ETD reduction for TFAS relative to outcome-only RL.}} \rev{Each point is one model--environment setting; the dashed line is a least-squares fit.} \rev{The two quantities have the same rank order across all six settings.}}
    \label{fig:etd_vs_accuracy}
    \vspace{-0.1cm}
\end{figure}

\subsection{\rev{Post-Training Failure-Context Accumulation}}
\label{subsec:post_training_etd}

\rev{The ETD column of Table~\ref{tab:main_results} tests whether execution-validity training changes the context quantity identified in Section~\ref{sec:cost_of_failure}.} \rev{We compare the base model and the three trained policies within each environment and model scale.} \rev{The reported statistic is mean final ETD over all evaluation episodes, with zero assigned to histories without labeled failure spans.}

\rev{TFAS has the lowest ETD among the four strategies in all six settings.} \rev{Relative to outcome-only RL, ETD decreases from 1.770\% to 0.923\% on ReTool-4B and from 0.830\% to 0.610\% on ReTool-8B.} \rev{On OpenThinkImg, the corresponding changes are 3.420\% to 1.510\% for the 4B model and 1.106\% to 0.801\% for the 8B model.} \rev{On ALFWorld, ETD decreases from 23.700\% to 10.070\% for Qwen3-4B and from 8.720\% to 2.820\% for Qwen3-8B, reductions of 13.630 and 5.900 percentage points.} \rev{Across the six settings, the relative ETD reduction over outcome-only RL ranges from 26.5\% to 67.7\%.} \rev{TFAS also has lower ETD than the untrained base model in every setting.}

\rev{Reward-level validity shaping does not produce the same pattern.} \rev{Tool Valid Reward increases ETD over outcome-only RL in both ReTool settings and ALFWorld-4B, while it decreases ETD in both OpenThinkImg settings and ALFWorld-8B.} \rev{For example, ALFWorld-4B reaches 36.110\% under Tool Valid Reward, compared with 23.700\% under outcome-only RL and 10.070\% under TFAS.} \rev{Adding execution validity at the reward level therefore does not reliably reduce failure-context density, whereas mixing it at the advantage level does so in every setting.}

\rev{\noindent\textbf{Accuracy gains track ETD reductions.}} \rev{The two effects of TFAS are closely aligned across settings (Figure~\ref{fig:etd_vs_accuracy}).} \rev{Ranking the six settings by relative ETD reduction over outcome-only RL gives ALFWorld-8B (67.7\%), ALFWorld-4B (57.5\%), OpenThinkImg-4B (55.8\%), ReTool-4B (47.9\%), OpenThinkImg-8B (27.6\%), and ReTool-8B (26.5\%); the corresponding gains in Overall Acc are 10.0, 6.5, 5.0, 4.3, 3.7, and 3.3 points, in exactly the same order.} \rev{The settings in which TFAS removes the most failure context are those in which it improves task completion the most, consistent with the cost of retained failures identified in Section~\ref{sec:cost_of_failure}.}

\subsection{Ablation: Advantage Mixing Weight}
\label{sec:lambda_ablation}

Figure~\ref{fig:lambda_sensitivity} studies the advantage mixing weight $\lambda$ on OpenThinkImg with Qwen3-VL-4B~\cite{bai2025qwen3vl}, keeping the final accuracy reward, policy-optimization configuration, and training budget fixed. The $\lambda=0$ setting corresponds to outcome-only optimization, while larger values increase the contribution of tool-feedback advantage in Equation~\ref{eq:mixed_advantage}. Adding tool-use feedback improves overall task success and reduces error-containing trajectories, with $\lambda=0.5$ giving the best balance in this setting. When $\lambda$ becomes too large, the tool signal can dominate the global task objective and reduce final accuracy, indicating that tool-feedback advantage is most effective as a complementary shaping signal rather than a replacement for task-level credit. The best $\lambda$ may vary across environments and models; the values used in this paper are reported in Appendix~\ref{apdx:hyper}.

\subsection{Qualitative Analysis}

Figure~\ref{fig:qualitative_tool_failure_recovery} shows the ALFWorld case behind the gains in Table~\ref{tab:main_results}. The baseline reaches the desk, takes the bowl, then attempts an unavailable action; the resulting error feedback triggers repeated invalid actions. After TFAS training, the agent follows the executable layout and completes the task in 6 steps without invalid actions. This case illustrates how TFAS favors trajectories with higher execution validity, reducing the introduction of invalid actions and their associated error feedback. Similar examples appear in Appendix~\ref{apdx:qualitative}.


\FloatBarrier

\Needspace{16\baselineskip}
\section{Conclusion}
\label{sec:conclusion}

\rev{An agent's failed interactions remain in the history that conditions its later decisions.} \rev{We measure this accumulated failure context with ETD, and matched ALFWorld interventions show that its content lowers continuation success.} \rev{Failure-conditioned evaluations show the same gap across planning, code execution, and visual tool use.} \rev{TFAS, a trajectory-level execution-validity signal, improves overall accuracy and achieves the lowest ETD in all six settings, reducing ETD by 26.5--67.7\% relative to outcome-only RL.} \rev{Across these settings, larger ETD reductions come with larger accuracy gains, in exactly the same rank order.} \rev{Agents should therefore be trained and evaluated on both task completion and the failure content they leave in their history.}

\bibliographystyle{ieeenat_fullname}
{\small\bibliography{main}}
\clearpage
\appendix

\section{Training Hyperparameters} \label{apdx:hyper}
In this section, Tables~\ref{tab:hyper_retool_4b}--\ref{tab:hyper_alfworld_8b} report the hyperparameter settings used for all reinforcement-learning experiments in Section~\ref{sec:experiment}. We list the configurations separately for each environment and model size, covering ReTool, OpenThinkImg, and ALFWorld. Within each table, the three columns correspond to the outcome-only reward baseline, the tool-valid reward baseline, and our Tool-Feedback Advantage Shaping (TFAS) strategy, respectively. Unless otherwise specified, hyperparameters are kept identical across reward strategies within the same environment-model setting to ensure a controlled comparison. We use GRPO~\cite{shao2024deepseekmath} and DAPO~\cite{yu2025dapo} as the underlying PPO-style optimizers.

 \section{Method Details} \label{apdx:method}
\subsection{Core Codes for TFAS}

In this section, we show the core codes of TFAS:
\begin{lstlisting}[breakautoindent=false,breakindent=1em,caption={Core code for Tool-Feedback Advantage Shaping.}, label={lst:tfas}]
"""
Dual Advantage for Multi-Turn GRPO

Key idea:
    final_adv = (1 - w) * success_advantage + w * valid_fraction_advantage

Instead of mixing two reward signals into one scalar before GRPO normalisation,
we normalise each signal independently and blend the resulting advantages.
This preserves gradient signal when the success reward has zero variance
(all-fail or all-succeed prompt groups).
"""

from __future__ import annotations
from collections import defaultdict

import numpy as np
import torch


def _grpo_normalise(scores: torch.Tensor, group_ids: list[str],
                    eps: float = 1e-6) -> torch.Tensor:
    """Within-group zero-mean, unit-variance normalisation (standard GRPO).
    Groups with zero variance get advantage = 0."""
    adv = torch.empty_like(scores)
    groups: dict[str, list[int]] = defaultdict(list)
    for i, gid in enumerate(group_ids):
        groups[gid].append(i)
    for idxs in groups.values():
        g = scores[idxs]
        mu, std = g.mean(), g.std() if len(idxs) > 1 else torch.tensor(1.0)
        adv[idxs] = (g - mu) / (std + eps) if std > eps else 0.0
    return adv


def compute_dual_advantage(
    rm_scores: torch.Tensor,       # [N, seq_len] success reward (0/1 at last token)
    response_mask: torch.Tensor,   # [N, seq_len] response token mask
    prompt_uids: np.ndarray,       # [N] prompt group id per row
    traj_uids: np.ndarray,         # [N] trajectory id per row
    is_valid_action: np.ndarray,   # [N] 1 if step action was valid
    valid_fraction_coef: float,    # coefficient to scale valid-fraction score
    vf_adv_weight: float,          # blend weight w (e.g. 0.5)
    eps: float = 1e-6,
) -> tuple[torch.Tensor, dict]:
    """Compute dual advantage with separate GRPO normalisation per signal."""

    # -- Step 1: aggregate per-step rows into per-trajectory scalars -----
    row_scores = rm_scores.sum(dim=-1)  # [N]

    traj_to_rows: dict[str, list[int]] = defaultdict(list)
    for i, uid in enumerate(traj_uids):
        traj_to_rows[str(uid)].append(i)

    traj_ids = list(traj_to_rows.keys())
    success_scores, vf_scores, traj_prompts = [], [], []

    for tid in traj_ids:
        rows = traj_to_rows[tid]
        # Success: same binary reward across all steps in a trajectory
        success_scores.append(row_scores[rows[0]])
        # Valid-fraction: proportion of valid actions, scaled by coefficient
        vf = float(is_valid_action[rows].sum()) / max(len(rows), 1)
        vf_scores.append(torch.tensor(vf * valid_fraction_coef,
                                      device=rm_scores.device))
        traj_prompts.append(str(prompt_uids[rows[0]]))

    success_t = torch.stack(success_scores)
    vf_t = torch.stack(vf_scores)

    # -- Step 2: GRPO normalise each signal independently -----
    # This is the core of dual advantage: each reward signal gets its own
    # within-group normalisation, so neither signal can drown out the other.
    success_adv = _grpo_normalise(success_t, traj_prompts, eps)
    vf_adv = _grpo_normalise(vf_t, traj_prompts, eps)

    # -- Step 3: blend -----
    blended = (1.0 - vf_adv_weight) * success_adv + vf_adv_weight * vf_adv

    # -- Step 4: broadcast back to per-step rows -----
    advantages = torch.zeros_like(response_mask, dtype=torch.float32)
    for idx, tid in enumerate(traj_ids):
        rows = traj_to_rows[tid]
        advantages[rows] = blended[idx] * response_mask[rows]

    metrics = {
        "dual_adv/success_adv_mean": float(success_adv.mean()),
        "dual_adv/vf_adv_mean": float(vf_adv.mean()),
        "dual_adv/blended_adv_mean": float(blended.mean()),
    }
    return advantages, metrics

\end{lstlisting}

\begin{table*}[!t]
    \centering
    \caption{\textbf{Hyperparameter settings for Qwen3-4B ReTool experiments.}}
    \renewcommand\arraystretch{1.3}
    \resizebox{\textwidth}{!}{%
    \begin{tabular}{c | ccc}
    \hline
        ~ & Outcome Reward & Tool Valid Reward & TFAS (Ours) \\ \hline \hline
        Pretrained Model & Qwen3-4B-Instruct-2507~\cite{qwen3technicalreport} & Qwen3-4B-Instruct-2507~\cite{qwen3technicalreport} & Qwen3-4B-Instruct-2507~\cite{qwen3technicalreport} \\ 
        RL Algorithm & DAPO & DAPO & DAPO \\ 
        Train Batchsize & 64 & 64 & 64 \\ 
        Temperature & 1.0 & 1.0 & 1.0 \\ 
        Num Response per Prompt & 16 & 16 & 16 \\ 
        Max Prompt Length & 2048 & 2048 & 2048 \\ 
        Max Response Length & 16384 & 16384 & 16384 \\ 
        Bf16 & True & True & True \\ 
        Actor Learning Rate & 1e-6 & 1e-6 & 1e-6 \\ 
        KL Loss Coef & 0.0 & 0.0 & 0.0 \\ 
        Steps & 40 & 40 & 40 \\
        $\lambda$ & -- & 0.5 & 0.5 \\
        \hline
    \end{tabular}%
    }
    \label{tab:hyper_retool_4b}
    \end{table*}

\begin{table*}[!t]

    \centering
    \caption{\textbf{Hyperparameter settings for Qwen3-8B ReTool experiments.}}
    \renewcommand\arraystretch{1.3}
    \resizebox{\textwidth}{!}{%
    \begin{tabular}{c | ccc}
    \hline
        ~ & Outcome Reward & Tool Valid Reward & TFAS (Ours) \\ \hline \hline
        Pretrained Model & Qwen3-8B~\cite{qwen3technicalreport} & Qwen3-8B~\cite{qwen3technicalreport} & Qwen3-8B~\cite{qwen3technicalreport} \\ 
        RL Algorithm & DAPO & DAPO & DAPO \\ 
        Train Batchsize & 64 & 64 & 64 \\ 
        Temperature & 1.0 & 1.0 & 1.0 \\ 
        Num Response per Prompt & 16 & 16 & 16 \\ 
        Max Prompt Length & 2048 & 2048 & 2048 \\ 
        Max Response Length & 16384 & 16384 & 16384 \\ 
        Bf16 & True & True & True \\ 
        Actor Learning Rate & 1e-6 & 1e-6 & 1e-6 \\ 
        KL Loss Coef & 0.0 & 0.0 & 0.0 \\ 
        Steps & 40 & 40 & 40 \\
        $\lambda$ & -- & 0.5 & 0.5 \\
        \hline
    \end{tabular}%
    }
    \label{tab:hyper_retool_8b}
    \end{table*}

\begin{table*}[!t]

    \centering
    \caption{\textbf{Hyperparameter settings for Qwen3-VL-4B OpenThinkImg experiments.}}
    \renewcommand\arraystretch{1.3}
    \resizebox{\textwidth}{!}{%
    \begin{tabular}{c | ccc}
    \hline
        ~ & Outcome Reward & Tool Valid Reward & TFAS (Ours) \\ \hline \hline
        Pretrained Model & Qwen3-VL-4B-Instruct~\cite{bai2025qwen3vl} & Qwen3-VL-4B-Instruct~\cite{bai2025qwen3vl} & Qwen3-VL-4B-Instruct~\cite{bai2025qwen3vl} \\ 
        RL Algorithm & GRPO & GRPO & GRPO \\ 
        Train Batchsize & 64 & 64 & 64 \\ 
        Temperature & 1.0 & 1.0 & 1.0 \\ 
        Num Response per Prompt & 8 & 8 & 8 \\ 
        Max Prompt Length & 4096 & 4096 & 4096 \\ 
        Max Response Length & 512 & 512 & 512 \\ 
        Bf16 & True & True & True \\ 
        Actor Learning Rate & 1e-6 & 1e-6 & 1e-6 \\ 
        KL Loss Coef & 0.0 & 0.0 & 0.0 \\ 
        Steps & 8 & 8 & 8 \\
        $\lambda$ & -- & 0.5 & 0.5 \\
        \hline
    \end{tabular}%
    }
    \label{tab:hyper_openthinkimg_4b}
\end{table*}

\begin{table*}[!t]
    \centering
    \caption{\textbf{Hyperparameter settings for Qwen3-VL-8B OpenThinkImg experiments.}}
    \renewcommand\arraystretch{1.3}
    \resizebox{\textwidth}{!}{%
    \begin{tabular}{c | ccc}
    \hline
        ~ & Outcome Reward & Tool Valid Reward & TFAS (Ours) \\ \hline \hline
        Pretrained Model & Qwen3-VL-8B-Instruct~\cite{bai2025qwen3vl} & Qwen3-VL-8B-Instruct~\cite{bai2025qwen3vl} & Qwen3-VL-8B-Instruct~\cite{bai2025qwen3vl} \\ 
        RL Algorithm & GRPO & GRPO & GRPO \\ 
        Train Batchsize & 16 & 16 & 16 \\ 
        Temperature & 1.0 & 1.0 & 1.0 \\ 
        Num Response per Prompt & 8 & 8 & 8 \\ 
        Max Prompt Length & 4096 & 4096 & 4096 \\ 
        Max Response Length & 512 & 512 & 512 \\ 
        Bf16 & True & True & True \\ 
        Actor Learning Rate & 1e-6 & 1e-6 & 1e-6 \\ 
        KL Loss Coef & 0.0 & 0.0 & 0.0 \\ 
        Steps & 8 & 8 & 8 \\
        $\lambda$ & -- & 0.5 & 0.5 \\
        \hline
    \end{tabular}%
    }
    \label{tab:hyper_openthinkimg_8b}
    \end{table*}

\begin{table*}[!t]

    \centering
    \caption{\textbf{Hyperparameter settings for Qwen3-4B ALFWorld experiments.}}
    \renewcommand\arraystretch{1.3}
    \resizebox{\textwidth}{!}{%
    \begin{tabular}{c | ccc}
    \hline
        ~ & Outcome Reward & Tool Valid Reward & TFAS (Ours) \\ \hline \hline
        Pretrained Model & Qwen3-4B-Instruct-2507~\cite{qwen3technicalreport} & Qwen3-4B-Instruct-2507~\cite{qwen3technicalreport} & Qwen3-4B-Instruct-2507~\cite{qwen3technicalreport} \\ 
        RL Algorithm & GRPO & GRPO & GRPO \\ 
        Train Batchsize & 64 & 64 & 64 \\ 
        Temperature & 1.0 & 1.0 & 1.0 \\ 
        Num Response per Prompt & 8 & 8 & 8 \\ 
        Max Prompt Length & 12000 & 12000 & 12000 \\ 
        Max Response Length & 256 & 256 &  256 \\ 
        Bf16 & True & True & True \\ 
        Actor Learning Rate & 1e-6 & 1e-6 & 1e-6 \\ 
        KL Loss Coef & 0.15 & 0.15 & 0.15 \\ 
        Steps & 23 & 23 & 23 \\
        $\lambda$ & -- & 0.17 & 0.17 \\
        \hline
    \end{tabular}%
    }
    \label{tab:hyper_alfworld_4b}
    \end{table*}

\begin{table*}[!t]

    \centering
    \caption{\textbf{Hyperparameter settings for Qwen3-8B ALFWorld experiments.}}
    \renewcommand\arraystretch{1.3}
    \resizebox{\textwidth}{!}{%
    \begin{tabular}{c | ccc}
    \hline
        ~ & Outcome Reward & Tool Valid Reward & TFAS (Ours) \\ \hline \hline
        Pretrained Model & Qwen3-8B~\cite{qwen3technicalreport} & Qwen3-8B~\cite{qwen3technicalreport} & Qwen3-8B~\cite{qwen3technicalreport} \\ 
        RL Algorithm & GRPO & GRPO & GRPO \\ 
        Train Batchsize & 64 & 64 & 64 \\ 
        Temperature & 1.0 & 1.0 & 1.0 \\ 
        Num Response per Prompt & 16 & 16 & 16 \\ 
        Max Prompt Length & 12000 & 12000 & 12000 \\ 
        Max Response Length & 256 & 256 & 256 \\ 
        Bf16 & True & True & True \\ 
        Actor Learning Rate & 1e-6 & 1e-6 & 1e-6 \\ 
        KL Loss Coef & 0.15 & 0.15 & 0.15 \\ 
        Steps & 23 & 23 & 23 \\
        $\lambda$ & -- & 0.17 & 0.17 \\
        \hline
    \end{tabular}%
    }
    \label{tab:hyper_alfworld_8b}
\end{table*}

\clearpage

\begin{table*}[t]
    \centering
    \caption{\textbf{Training and evaluation data used for each environment.}}
    \renewcommand\arraystretch{1.25}
    \resizebox{\textwidth}{!}{%
    \begin{tabular}{p{0.18\textwidth} p{0.38\textwidth} p{0.38\textwidth}}
    \toprule
    \textbf{Environment} & \textbf{Training Set} & \textbf{Test / Evaluation Set} \\
    \midrule
    ReTool & DAPO-Math-17k~\cite{yu2025dapo} & AIME 2025~\cite{maa2025aime} \\
    OpenThinkImg & OpenThinkIMG-Chart-RL-14501~\cite{su2025openthinkimg} & CharXiv~\cite{wang2024charxiv} \\
    ALFWorld & ALFWorld~\cite{shridhar2021alfworld} & ALFWorld~\cite{shridhar2021alfworld} \\
    \bottomrule
    \end{tabular}%
    }
    \label{tab:dataset_splits}
    \vspace{1.5em}

    \centering
    \caption{\textbf{Independent repeats and 95\% CIs for the ALFWorld main experiment.} ``Runs'' lists the three independent results, and the adjacent column reports the sample mean $\pm$ the two-sided 95\% CI half-width. All values are percentages. ETD Run 1 is copied from Table~\ref{tab:main_results}.}
    \label{tab:alfworld_ci}
    \renewcommand\arraystretch{1.18}
    \resizebox{\textwidth}{!}{%
    \begin{tabular}{ll cc cc cc cc}
    \toprule
    \multirow{2}{*}{\textbf{Model}} & \multirow{2}{*}{\textbf{Reward Strategy}} &
    \multicolumn{2}{c}{\textbf{Overall Acc ($\uparrow$)}} &
    \multicolumn{2}{c}{\textbf{Err Proportion ($\downarrow$)}} &
    \multicolumn{2}{c}{\textbf{Err Acc ($\uparrow$)}} &
    \multicolumn{2}{c}{\textbf{ETD ($\downarrow$)}} \\
    \cmidrule(lr){3-4} \cmidrule(lr){5-6} \cmidrule(lr){7-8} \cmidrule(lr){9-10}
    & & Runs 1/2/3 & Mean $\pm$ 95\% CI & Runs 1/2/3 & Mean $\pm$ 95\% CI & Runs 1/2/3 & Mean $\pm$ 95\% CI & Runs 1/2/3 & Mean $\pm$ 95\% CI \\
    \midrule
    \multirow{3}{*}{Qwen3-4B}
    & Original Outcome Reward  & $17.8 / 16.7 / 17.9$ & $17.5 \pm 1.65$ & $84.2 / 85.2 / 84.0 $ & $84.5 \pm 1.60$ & $12.3 / 12.0 / 12.0 $ & $12.1 \pm 0.43$ & $23.7 / 23.9 / 23.0 $ & $23.5 \pm 1.17$ \\
    & + Tool Valid Reward      & $20.9 / 21.3 / 21.0$ & $21.1 \pm 0.52$ & $66.7 / 65.5 / 66.7 $ & $66.3 \pm 1.72$ & $22.1 / 21.8 / 21.9$ & $21.9 \pm 0.38$ & $36.1 / 35.2 / 36.2 $ & $35.8 \pm 1.37$ \\
    & + TFAS (Ours)            & $24.3 / 24.5 / 23.9$ & $24.2 \pm 0.76$ & $64.1 / 63.5 / 64.5$ & $64.0 \pm 1.25$ & $31.8 / 31.9 / 31.5$ & $31.7 \pm 0.52$ & $10.1 / 11.3 / 9.7 $ & $10.4 \pm 2.07$ \\
    \midrule
    \multirow{3}{*}{Qwen3-8B}
    & Original Outcome Reward  & $31.2 / 30.0 / 32.4$ & $31.2 \pm 2.98$ & $18.3 / 18.9 / 17.3$ & $18.2 \pm 2.01$ & $21.0 / 20.1 / 22.5$ & $21.2 \pm 3.01$ & $8.7 / 8.0 / 9.0 $ & $8.6 \pm 1.27$ \\
    & + Tool Valid Reward      & $33.6 / 33.4 / 33.0$ & $33.3 \pm 0.76$ & $18.0 / 19.8 / 18.7$ & $18.8 \pm 2.25$ & $35.7 / 35.4 / 35.3$ & $35.5 \pm 0.52$ & $8.0 / 8.8 / 8.1 $ & $8.3 \pm 1.08$ \\
    & + TFAS (Ours)            & $41.2 / 42.1 / 40.8$ & $41.4 \pm 1.65$ & $15.8 / 14.8 / 16.5$ & $15.7 \pm 2.12$ & $52.4 / 52.3 / 52.3$ & $52.3 \pm 0.14$ & $2.8 / 2.5 / 2.5 $ & $2.6 \pm 0.43$ \\
    \bottomrule
    \end{tabular}%
    }
\end{table*}

\subsection{Other Details}
 We implement our method based on the \textbf{verl} reinforcement learning framework, into which we have integrated the aforementioned ALFWorld, ReTool, and OpenThinkImg environments. To facilitate efficient distributed training, we utilize the \textbf{FSDP} (Fully Sharded Data Parallel) backend for all experiments. The base models include Qwen3-4B~\cite{qwen3technicalreport}, Qwen3-8B~\cite{qwen3technicalreport}, Qwen3-VL-4B~\cite{bai2025qwen3vl}, and Qwen3-VL-8B~\cite{bai2025qwen3vl}. Experiments were conducted on A800 GPUs; the exact hardware allocation is reported in Appendix~\ref{apdx:compute_resources}. Detailed hyperparameters and other implementation configurations are provided in Appendix~\ref{apdx:hyper}.

\section{Experimental Details} \label{apdx:exp_details}

Table~\ref{tab:dataset_splits} summarizes the training and evaluation data used in the three environments. ReTool uses the public RL training and validation sets released with the ReTool project~\cite{feng2025retoolreinforcementlearningstrategic}, OpenThinkImg uses the chart-reasoning RL and held-out test sets~\cite{su2025openthinkimg}, and ALFWorld follows the standard environment split for training and held-out evaluation~\cite{shridhar2021alfworld}.


\section{ETD Analysis Details} \label{apdx:etd_analysis}
 
%


We evaluate Qwen3-8B on the ALFWorld TextWorld environment~\cite{shridhar2021alfworld}. Each game is rolled out four times. The agent is served with vLLM ; decoding is greedy ($T=0$, top-$p=1.0$) with at most $256$ new tokens per turn and a repetition penalty of $1.05$, so every trajectory is
reproducible from the seed. An episode terminates when the environment signals
task completion or when the budget of $B=20$ turns is exhausted.

\section{Independent Runs and Confidence Intervals}
\label{apdx:confidence_intervals}

To assess the variability of the reported results, we conduct three independent runs for each configuration. For trained configurations, each run includes both training and evaluation. For the untrained base models, only evaluation is repeated under the same protocol. Across runs, we keep the datasets, model configurations, hyperparameters, and evaluation metrics unchanged. We report the mean over the three runs together with its two-sided 95\% confidence interval (CI).

\subsection{Independent Repeats for the ALFWorld Main Experiment}

Table~\ref{tab:alfworld_ci} reports the independent repeats for the ALFWorld portion of the main experiment. Across both model scales, TFAS obtains the highest mean Overall Acc and Err Acc among the compared reward strategies.

\subsection{Independent Repeats for the Controlled Context Interventions}
\label{apdx:intervention_repeats}

We repeat the controlled ALFWorld interventions at $\rho=0.25$ from Section~\ref{subsec:controlled_interventions} three times. Each repeat reruns every matched branch with an independent evaluation seed while keeping the evaluated tasks, anchor construction, intervention content, and remaining action budget fixed. Table~\ref{tab:intervention_ci} reports the change in success rate relative to the corresponding matched trajectory with an unmodified history.

\begin{table}[t]
    \centering
    \caption{\textbf{Independent repeats and 95\% CIs for the controlled failure-context interventions on ALFWorld at $\rho=0.25$.} Values are changes in success rate (percentage points); $E_k$, $N_k$, and $D_k$ denote error insertion, token-matched neutral insertion, and failure deletion, respectively.}
    \label{tab:intervention_ci}
    \renewcommand\arraystretch{1.16}
    \resizebox{\columnwidth}{!}{%
    \begin{tabular}{lc cc}
    \toprule
    \textbf{Intervention} & $\boldsymbol{k}$ & Runs 1/2/3 & Mean $\pm$ 95\% CI \\
    \midrule
    \multirow{3}{*}{Error insertion ($E_k$)}
    & 1 & $-4.7 / -4.2 / -4.9 $ & $-4.6 \pm 0.90$ \\
    & 2 & $-5.7 / -5.7 / -5.4$ & $-5.6 \pm 0.43$ \\
    & 3 & $-8.2 / -8.0 / -8.0$ & $-8.1 \pm 0.29$ \\
    \midrule
    \multirow{3}{*}{Neutral insertion ($N_k$)}
    & 1 & $-0.9 / -1.2 / -0.7$ & $-0.9 \pm 0.63$ \\
    & 2 & $-3.8 / -3.6 / -3.6$ & $-3.7 \pm 0.29$ \\
    & 3 & $-5.6 / -5.4 / -5.6$ & $-5.5 \pm 0.29$ \\
    \midrule
    \multirow{2}{*}{Failure deletion ($D_k$)}
    & 1 & $+2.5 / +2.7 / +2.7$ & $+2.6 \pm 0.29$ \\
    & 2 & $+20.0 / +19.8 / +20.3$ & $+20.0 \pm 0.63$ \\
    \bottomrule
    \end{tabular}%
    }
\end{table}
\FloatBarrier

\section{Qualitative Analysis} \label{apdx:qualitative}
\begin{figure*}[tp]
    \centering
    \includegraphics[width=\textwidth,height=0.78\textheight,keepaspectratio]{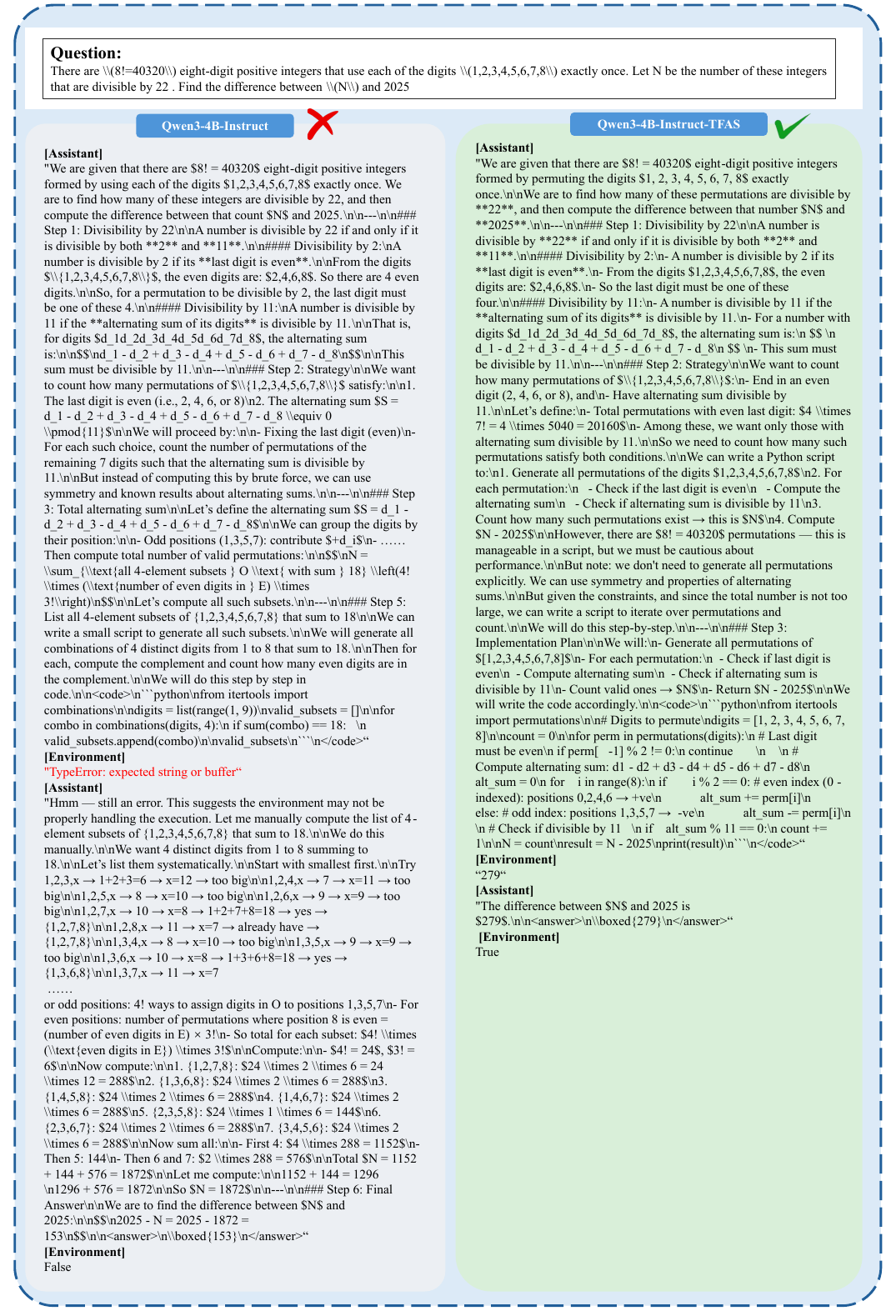}
    \caption{\textbf{Qualitative example on ReTool.} The case shows how TFAS affects tool-augmented mathematical reasoning behavior.}
    \label{fig:appendix_retool_qualitative}
\end{figure*}

\begin{figure*}[tp]
    \centering
    \includegraphics[width=\textwidth,height=0.78\textheight,keepaspectratio]{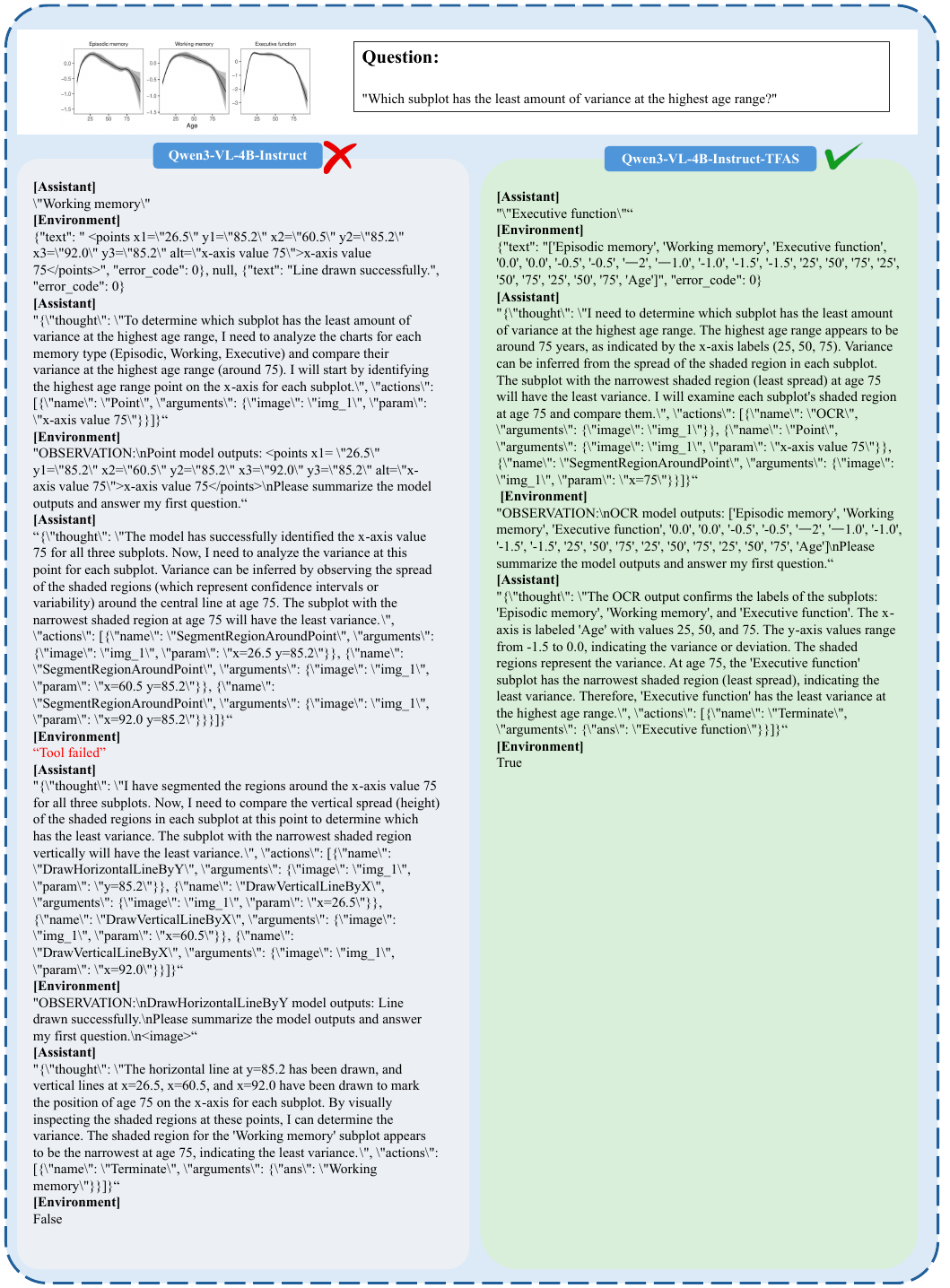}
    \caption{\textbf{Qualitative example on OpenThinkImg.} The case shows how TFAS affects tool-augmented visual reasoning behavior.}
    \label{fig:appendix_openimg_qualitative}
\end{figure*}

\begin{figure*}[tp]
    \centering
    \includegraphics[width=\textwidth,height=0.78\textheight,keepaspectratio]{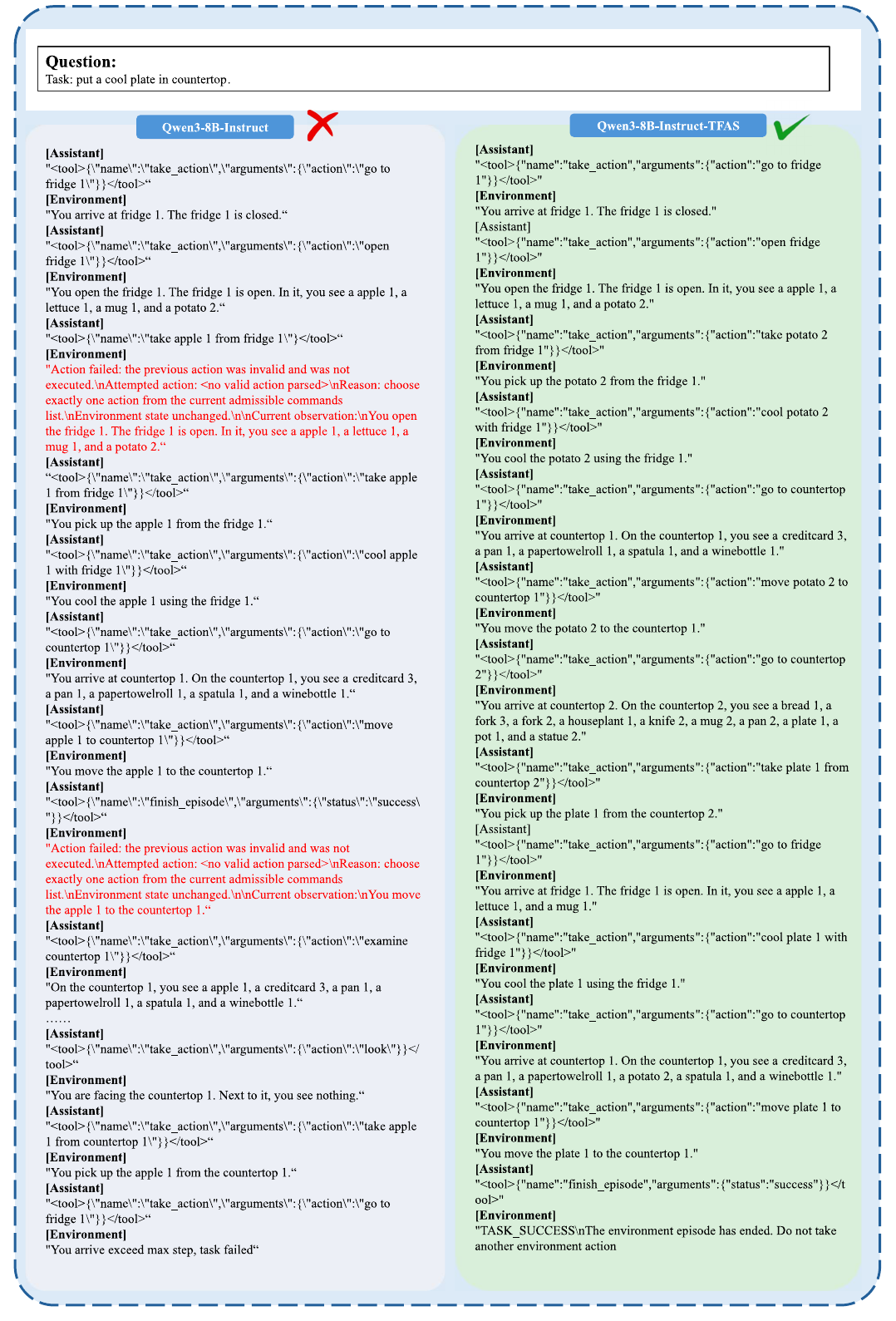}
    \caption{\textbf{Qualitative example on ALFWorld.} The case shows how TFAS affects embodied planning behavior.}
    \label{fig:appendix_alfworld_qualitative}
\end{figure*}

This section provides additional qualitative examples from the three evaluation environments. Figure~\ref{fig:appendix_retool_qualitative}, Figure~\ref{fig:appendix_openimg_qualitative}, and Figure~\ref{fig:appendix_alfworld_qualitative} show representative cases from ReTool, OpenThinkImg, and ALFWorld, respectively. Together, these cases illustrate how TFAS changes agent behavior across code/tool-use reasoning, visual reasoning, and embodied planning.

\section{Compute Resources} \label{apdx:compute_resources}
All reinforcement-learning experiments were conducted on NVIDIA A800 GPUs with Ubuntu 22.04, CUDA 12.6, bfloat16 training, and PyTorch Fully Sharded Data Parallel (FSDP)~\cite{zhao2023pytorchfsdp}. The OpenThinkImg experiments use 8 A800 GPUs per run for 6 hours. The remaining experiments, including all ReTool and ALFWorld runs, use 4 A800 GPUs per run. A ReTool training run takes approximately 50 hours, while an ALFWorld training run takes approximately 45 hours under the reported configuration. The training budgets, batch sizes, number of responses per prompt, maximum prompt and response lengths, and optimization steps are reported in Tables~\ref{tab:hyper_retool_4b}--\ref{tab:hyper_alfworld_8b}.

\section{Limitations} \label{apdx:limitation}
The method introduces a mixing weight $\lambda$, which controls the strength of tool-feedback advantage relative to task-accuracy advantage. Although our sensitivity study shows that a moderate value works well, this hyperparameter may still need to be tuned for different environments, models, or tool-feedback definitions. TFAS also adds a small amount of training-time computation because it estimates and normalizes a separate trajectory-level tool-feedback advantage before policy optimization. This overhead is minor compared with rollout generation and model updates in our experiments, but it is still an additional component beyond outcome-only RL.

\section{Broader Impact} \label{apdx:broad_impact}
This work aims to improve the reliability of long-horizon language-model agents by reducing invalid tool calls, execution errors, and other forms of failure-induced context pollution. Potential positive impacts include more dependable agent behavior in tool-assisted reasoning, visual analysis, and embodied planning settings, especially when repeated trial-and-error interactions would otherwise create avoidable failures.

As with general methods for improving agent capabilities, downstream applications should still consider the reliability and oversight requirements of the target domain. Better tool-use behavior may reduce avoidable execution errors, but it does not by itself guarantee that an agent's goals, tool permissions, or deployment context are appropriate. Our experiments are limited to benchmark environments and do not introduce new user data, scraped datasets, deployed systems, or released high-risk models. We therefore view the direct societal risk of this work as limited, while recommending that practical deployments use task-appropriate safeguards such as tool permission controls and human review for consequential decisions.


\section{Other}
We use the following existing assets and cite their original sources in the manuscript. All assets are used for research evaluation and training under their respective usage terms.

\begin{itemize}
    \item \textbf{Benchmarks and datasets.} The experiments use ReTool~\cite{feng2025retoolreinforcementlearningstrategic}, OpenThinkImg~\cite{su2025openthinkimg}, and ALFWorld~\cite{shridhar2021alfworld} as the main interactive environments. The corresponding training and evaluation sets are DAPO-Math-17k~\cite{yu2025dapo} and AIME 2025~\cite{maa2025aime} for ReTool, OpenThinkIMG-Chart-RL-14501~\cite{su2025openthinkimg} and CharXiv~\cite{wang2024charxiv} for OpenThinkImg, and the standard ALFWorld splits~\cite{shridhar2021alfworld} for embodied planning.
    \item \textbf{Models.} The trainable base models are Qwen3-4B-Instruct-2507, Qwen3-8B~\cite{qwen3technicalreport}, Qwen3-VL-4B-Instruct, and Qwen3-VL-8B-Instruct~\cite{bai2025qwen3vl}. Additional closed-source models are used only for failure-conditioned evaluation in Section~\ref{sec:cost_of_failure}, following their public API terms.
    \item \textbf{Training and evaluation frameworks.} The reinforcement-learning pipeline is implemented with verl~\cite{verlproject2024verl} and PyTorch FSDP~\cite{zhao2023pytorchfsdp}. Model loading and evaluation use standard open-source libraries, including Transformers~\cite{wolf2020transformers}.
\end{itemize}

\end{document}